\documentclass[conference]{IEEEtran}
\IEEEoverridecommandlockouts
\usepackage{cite}
\usepackage{amsmath,amssymb,amsfonts}
\usepackage{algorithmic}
\usepackage{graphicx}
\usepackage{textcomp}
\usepackage{xcolor}
\usepackage{multirow}
\def\BibTeX{{\rm B\kern-.05em{\sc i\kern-.025em b}\kern-.08em
    T\kern-.1667em\lower.7ex\hbox{E}\kern-.125emX}}
\begin{document}

\title{Improved Touchless Respiratory Rate Sensing\\
\thanks{\copyright 2022 IEEE. Personal use of this material is permitted. Permission from IEEE must be obtained for all other uses, in any current or future media, including reprinting/republishing this material for advertising or promotional purposes, creating new collective works, for resale or redistribution to servers or lists, or reuse of any copyrighted component of this work in other works.}
}

\author{\IEEEauthorblockN{Petro Franchuk}
\IEEEauthorblockA{\textit{SoftServe Inc.} \\
pfranc@softserveinc.com}
\and
\IEEEauthorblockN{Tetiana Yezerska}
\IEEEauthorblockA{\textit{Ukrainian Catholic University} \\
yezerska@ucu.edu.ua}
}

\maketitle

\begin{abstract}
Recently, remote respiratory rate measurement techniques gained much attention as they were developed to overcome the limitations of device-based classical methods and manual counting. Many approaches for RR extraction from the video stream of the visible light camera were proposed, including the pixel intensity changes method. In this paper, we propose a new method for 1D profile creation for pixel intensity changes-based method, which significantly increases the algorithm’s performance. Additional accuracy gain is obtained via a new method of motion signals grouping presented in this work. We introduce several changes to the standard pipeline, which enables real-time continuous RR monitoring and allows applications in the human-computer interaction systems. Evaluation results on two internal and one public datasets showed 0.7 BPM, 0.6 BPM, and 1.4 BPM MAE, respectively.     
\end{abstract}


\section{Introduction}
Respiratory rate (RR), also known as breathing rate, is one of the most important human vital signs. Abnormal RR can be a sign of multiple diseases. According to \cite{1}, RR is sensitive to different pathological conditions such as adverse cardiac events, pneumonia, clinical deterioration, and stressors, including emotional stress, cognitive load, heat, cold, physical effort, and exercise-induced fatigue. Such a variety of tracked conditions makes RR an important biosignal for many medical and human-computer interaction (HCI) scenarios.      

For respiratory rate monitoring, such wearables as chest band, facemask device, spirometer, and oral-nasal cannula are usually used. However, such devices may cause discomfort for the patient wearing them for an extended period of time. Indeed, such wearables as smartwatches are more comfortable for long-term usage, but their users bear the extra cost of purchasing.

Going beyond just medical and wellness use-cases, touchless vital signs measurement opens broad opportunities for HCI systems development, allowing the development of real-time systems that analyze a person's health and emotional state without disturbing the user. In addition to facial expression data, vital signs such as respiratory and heart rate (HR) are excellent data sources to analyze the user's stress level and emotional state \cite{2, 3}. Potential applications range from improving social interactions\cite{4, 19} to novel types of interactive digital signage scenarios\cite{5}.  

Although many works have been published on remote RR measurement from visible-light cameras during the last decade, almost all of them either don't satisfy real-time performance requirement or don't provide a solution for continuous measurement.      

In this paper, we propose an improved pixel intensity changes (PIC) based RR measurement algorithm, which can be applied in a resting-state breathing use case. Our method involves several changes to the typical pipeline, which, to our knowledge, were not previously covered in the literature. Proposed improvements enable real-life applications, such as continuous real-time monitoring. The evaluation on two internal and one external dataset showed good performance (0.7 BPM, 0.6 BPM, and 1.4 BPM MAE, respectively). The main contribution of our work consists of:  
\begin{itemize}
\item  {a new method of 1D profile formation, which significantly improves the performance of the algorithm;}

\item{a new method for selecting and analysis of breathing motion signals, which outperforms previous methods;}

\item{an adaptation of PIC-based method for continuous real-time monitoring, which enables real-life applications, including HCI systems.}
\end{itemize}

\section{Related Works}

Most classical RGB camera-based approaches in the literature can be split into three categories: remote photoplethysmogram (rPPG) based, Optical Flow-based, and PIC-based. The rPPG-based methods require a very clean PPG signal, which is proven to be a hard task to accomplish for all popular vital signs such as HR, oxygen saturation (SpO2), and RR. Optical Flow and PIC-based methods show comparable performance, but the first one is much more computationally expensive due to the necessity to compute the shift on every frame. Hence, our method of choice is the pixel intensity changes-based approach, which should be compared to other PIC-based methods.

The general pipeline in most related PIC-based works \cite{6, 7, 8, 9, 10, 11} includes steps such as the detection of the region of interest (ROI), 1D profile creation, selection of the best motion signals and transformation it to respiratory wave (RW), RW post-processing and RR estimation either in time or frequency domains.  

Our method can also be compared to related works in terms of performance and applicability. Real-time performance is specified in such works as \cite{7, 11}, while in \cite{10}, in real-time only RW is measured, and RR is estimated offline. The continuous RR measurement is performed in such works as \cite{6, 8, 9, 10, 11}. 

\begin{figure*}[h!]
\centering
\includegraphics[width=\textwidth]{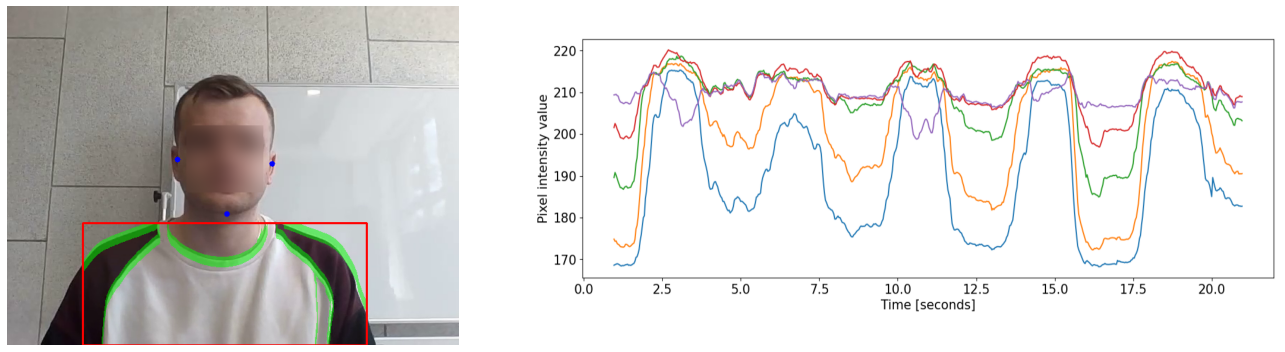}
\caption{Left: Example of ROI extraction. Blue points denote landmarks of the chin, left, and right ear; the red box denotes the chest region, from which motion is tracked; the green area indicates the detected edges, which will be used to create the 1D profile for this frame. Right: Example of the motion signals extracted from ROI (just part of all motion signals is visualized for better presentation).}
\label{fig:fig1}
\end{figure*}

\section{Our approach}
Our approach is based on the PIC method \cite{6,7, 8, 9, 10, 11}, which measures the RR by tracking the respiratory-induced motion of the human chest region. By observing the variation in the pixels’ intensity of this region, motion signals of the chest can be extracted and then converted to respiratory wave and respiratory rate. The proposed method works in real-time and can be applied for continuous RR monitoring, providing the RR value every second based on the last 20 seconds of data. Further, we describe each step of our approach in detail.

\subsection{ROI extraction}
A person's chest region is used as an ROI. The facial landmarks acquired from MediaPipe's\cite {12} FaceMesh solution are used to estimate the approximate shoulders location and ROI' dimensions. Figure\ref{fig:fig1}(left) shows an example of landmarks and ROI. The location and dimensions of the ROI are determined once per ROI re-estimation frequency, which, in our case, is 20 seconds.

\subsection{1D profile creation}
After ROI is detected and cropped, we need to transform this 2D (or, in the case of RGB image - 3D) tensor into a 1D vector that characterizes the ROI's color, texture, or structure. In literature, this 1D vector is referred to as a 1D profile and was for the first time described by Bartula et al.\cite{13}. To transform the image tensor to 1D profile, projection on a vertical axis (rows averaging) is usually used. In this work, such a 1D profile will be referred to as Full ROI. As far as we know, the only other 1D profile proposed is the work by Wang et al. \cite{14}, where the authors compared three different profiles (including classical row averaging). 

To create a Full ROI 1D profile, firstly, 3 RGB channels are averaged into a single value, after which each row is averaged into a single pixel, leading to a 1D vector (equal in size to the ROI height). The main drawback of such an approach is that during the row' averaging, a lot of noise and useless pixels are included in the final averaged value, which distorts the representation a lot.  

With these insights in mind, we propose a new method to create a 1D profile named Edges profile. So, before the channels' and rows' averaging, edges on the ROI are determined with the Canny edge detector\cite{15}. Figure\ref{fig:fig1}(left) shows an example of detected edges. Similar to the ROI's location and dimension, edge detection is done once per ROI re-estimation. Then, only edges and pixels around them (the dilation is also done to increase the area) are averaged and used in the final representation.   

Edges, especially horizontal, are good features for a 1D profile because they deviate more over time due to the vertical motion of the chest compared to non-edge pixels that stay the same or deviate little. To our knowledge, it is the first time edges are used in 1D profile creation. 

\subsection{Windowing \& ROI re-estimation}
1D profile vectors are stacked together into a 2D matrix. Each row in it corresponds to the respective row of the ROI and represents the changes in pixels intensity of that row over time. The motion signals are further processed using a moving window of 20 seconds, with a step size of 1 second. An example of such a window with motion signals is provided in Figure\ref{fig:fig1}(right). 

The ROI re-estimation is done every 20 seconds to adjust ROI absolute position on the image and its dimensions. In our approach, ROI re-estimation frequency is a subject of a trade-off between accuracy, latency, and robustness to the changes on the image.   

From one perspective, ROI re-estimation cannot be done too frequently since ROI will be moving along with the body, and almost no changes in pixel intensity will be visible. Besides this, ROI re-estimation is quite time-consuming because it includes face detection \& landmark prediction, along with edge detection. From another perspective, ROI re-estimation cannot be very rare because the situation in the video can significantly change (e.g., the person may move closer to the camera), and the algorithm should be able to adjust to the new location of the person`s chest without great delay. 

In the related works, RR is usually measured once per the whole recording. To our knowledge, there is just one work\cite{11}, in which authors explicitly claim continuous RR monitoring in real-time. The ability to provide continuous feedback in real-time and quickly adjust to new conditions allows our approach to be applied in such HCI applications as stress level estimation.

\subsection{Motion signals processing}
After the window accumulated enough data points for prediction, motion signals are processed. Firstly, signals are detrended to remove the low-frequency trend. Afterward, top 30\% of the signals are selected based on their standard deviation and grouped by 5\% into six groups. Finally, each group of signals is averaged, resulting in 6 candidate respiratory waves. To our knowledge, it is the first time such a selection and grouping process was proposed in the literature. 

In the related works, only the top 5\% of all the motion signals are usually selected and averaged into a single respiratory wave\cite{6, 8}. However, the top 5\% of the motion signals can sometimes contain very deviating but noisy waves, which either have no or minimal respiratory component. This high deviation noise is usually caused by the subject's motion, while a respiratory motion may be more periodic but has a lower amplitude. Hence, by taking a wider subset of motion signals and dividing them into several groups, we aim to have at least a few good respiratory waves that are not affected by high-amplitude noise. Additionally, we counted the share of cases when the best respiratory rate lies in the 5\% group vs. when they belong to the broader next 25\%. We found that 81\% of the time, the best RR belongs to a broader group of signals. Hence, such a method of motion signals processing can provide a more accurate RR than previous approaches.

\subsection{Time-domain respiratory rate estimation}
After the respiratory waves are extracted, the respiratory rate can be computed via the time-domain analysis. Firstly, RWs are smoothed by moving average to remove the high-frequency noise. Secondly, using a peak detector, all peaks are detected in the RW. Since the PIC-based algorithm doesn't determine the direction of motion but the pixel color changes, the peak can denote both full inhale and exhale. Finally, an instantaneous RR is computed as the inter-beat-interval (IBI) between the consecutive peaks. This instantaneous RR will be rather a signal than a single value since it is computed for each successive pair of peaks. 

\subsection{Best group selection}
The previous steps resulted in 6 respiratory waves and 6 instantaneous RR signals. The best group is determined by a respiratory rate signal with the lowest standard deviation because this signal will be obtained from the least noisy RW signal.

\subsection{RR value prediction}
Finally, after the best group selection, we will obtain just one instantaneous RR signal, which values will be averaged and returned as the predicted average RR value for the last 20 seconds. 

\section{Experimental setting}

\subsection{Datasets}
We have used two internal datasets (Internal\_mobile and Internal\_web) and one public dataset (COHFACE\cite{16}) for evaluation purposes.   

Internal\_mobile and Internal\_web datasets have pseudo ground truth, meaning there are no reference respiratory belt readings. Instead, subjects were asked to control their breathing at frequencies of 15 and 20 BPM. In both datasets, 15 subjects were recorded, 2 records per subject (one for 15 and one for 20BPM breathing frequency). In total, there are 30 videos (2 minutes each) for each of these datasets. For the Internal\_mobile, videos were recorded on the frontal camera of Google Pixel phone, while for Internal\_web on the laptop’s webcam. The video FPS in both datasets is 30.   

COHFACE\cite{16} dataset contains respiratory belt readings (RW) and video recording for 40 subjects, 4 records for each, and 160 videos (1 minute each) in total. The videos were recorded on the laptop’ webcam with 20 FPS. We manually computed RR ground truth from the provided RW readings to evaluate our approach. 

Since our internal datasets have no real ground truth, they were used to validate the hypothesis and to increase the data variability by introducing different subject's appearances and camera specifications. On the contrary, COHFACE was used as a typical dataset, which represents the real-life distribution and respiratory behavior.

\subsection{Metrics}
Metrics such as MAE and Success Rate (percentage of the predictions within some margins away from the ground truth) with bounds of 2 BPM were used for evaluation.

\section{Results}
To show the influence of the proposed 1D profile, our pipeline was tested with both Full ROI and Edges types of 1D profile creation. Additionally, selecting 5\% and 30\% of motion signals was compared to demonstrate the drawback of having a single group of motion signals. Finally, the performance of our algorithm was compared with other works\cite{17, 18}, where the evaluation was done on the COHFACE dataset. 

Table \ref{table:table1} shows that the algorithm version with 30\% of motion signals selected for further processing, the Edges 1D profile introduces a significant improvement in the accuracy on the Internal\_web and COHFACE datasets (~ -0.3 MAE and + 5\% of SR 2) while slightly decreasing the performance on the Internal\_mobile dataset (~ -1 SR 2). For a version with 5\%, the Edges 1D profile improves the performance on all datasets (~ -0.1 MAE and + 2\% of SR 2). Since all decisions are made mainly based on the COHFACE dataset, we can claim that the Edges 1D profile improves the performance of a proposed PIC-based method.

Table \ref{table:table1} also shows that selecting 30\% of motion signals instead of 5\% significantly increases the accuracy of both Full ROI and Edges profiles on all datasets. The best results are obtained by combining an Edge 1D profile and the selection of 30\% of motion signals.

\begin{table}[]
\caption{Performance of the proposed method with different variations of 1D profile and motion signals grouping}
\centering
\begin{tabular}{|c|c|c|c|c|}
\hline
Dataset                           & 1D profile                & Motion signals \% & MAE            & SR 2          \\ \hline
\multirow{4}{*}{Internal\_mobile} & \multirow{2}{*}{Full ROI} & 5\%               & 1.091          & 87.2          \\ \cline{3-5} 
                                  &                           & 30\%              & \textbf{0.704} & \textbf{94.1} \\ \cline{2-5} 
                                  & \multirow{2}{*}{Edges}    & 5\%               & 0.982          & 89.2          \\ \cline{3-5} 
                                  &                           & 30\%              & \textbf{0.699} & 92.9          \\ \hline
\multirow{4}{*}{Internal\_web}    & \multirow{2}{*}{Full ROI} & 5\%               & 0.982          & 89.9          \\ \cline{3-5} 
                                  &                           & 30\%              & 0.888          & 90.9          \\ \cline{2-5} 
                                  & \multirow{2}{*}{Edges}    & 5\%               & 0.872          & 91.4          \\ \cline{3-5} 
                                  &                           & 30\%              & \textbf{0.602} & \textbf{95.2} \\ \hline
\multirow{4}{*}{COHFACE}          & \multirow{2}{*}{Full ROI} & 5\%               & 1.943          & 72.0          \\ \cline{3-5} 
                                  &                           & 30\%              & 1.715          & 74.3          \\ \cline{2-5} 
                                  & \multirow{2}{*}{Edges}    & 5\%               & 1.753          & 74.6          \\ \cline{3-5} 
                                  &                           & 30\%              & \textbf{1.397} & \textbf{79.4} \\ \hline
\end{tabular}
\label{table:table1}
\end{table}

The comparison of a proposed method with the related methods\cite{17, 18}, which were also evaluated on the public COHFACE dataset, is provided in Table \ref{table:table2}. Although our methods differ in the architecture and evaluation procedure, we include this comparison since these methods are, to our knowledge, the only which were benchmarked on this dataset.

\begin{table}[h!]
\caption{Performance comparison of the proposed method with other related works }
\centering
\begin{tabular}{|c|c|}
\hline
Method                                  & MAE   \\ \hline
TS-DAN \cite{17} (Full)                 & 5.350 \\ \hline
MT-TS-DAN \cite{17} (Full)              & 5.728 \\ \hline
MT-CAN \cite{18}                        & 4.083 \\ \hline
MTTS-CAN \cite{18}                      & 2.482 \\ \hline
Ours                                    & \textbf{1.397} \\ \hline
\end{tabular}
\label{table:table2}
\end{table}

\section{Conclusions}
In this work, a new method for creating the 1D profile was proposed and proved to increase the accuracy of the PIC-based algorithm for remote RR extraction. Additional improvement was acquired by introducing a new motion signal selection and grouping method. Evaluation results on two internal and one public dataset showed significant improvement compared to previous methods. Our pipeline is suitable for continuous user behavior analysis, as it returns measurements in real-time, making it possible to decide about the user`s emotional state based on the dynamics of RR frequency change.

\section{Acknowledgments}
The authors would like to thank Volodymyr Karpiv for the initial discussion of the 1D profile and motion signals selection. Additionally, the authors would like to thank Oleh Menchyshyn for his helpful contribution to the discussions and paper review.

Special thanks go to the Ukrainian Armed Forces, which by defending our country, allowed us to work on this publication.

\bibliographystyle{IEEEtran}
\bibliography{IEEEabrv, my_bibliography}

\begin{thebibliography}{10}
\providecommand{\url}[1]{#1}
\csname url@samestyle\endcsname
\providecommand{\newblock}{\relax}
\providecommand{\bibinfo}[2]{#2}
\providecommand{\BIBentrySTDinterwordspacing}{\spaceskip=0pt\relax}
\providecommand{\BIBentryALTinterwordstretchfactor}{4}
\providecommand{\BIBentryALTinterwordspacing}{\spaceskip=\fontdimen2\font plus
\BIBentryALTinterwordstretchfactor\fontdimen3\font minus
  \fontdimen4\font\relax}
\providecommand{\BIBforeignlanguage}[2]{{%
\expandafter\ifx\csname l@#1\endcsname\relax
\typeout{** WARNING: IEEEtran.bst: No hyphenation pattern has been}%
\typeout{** loaded for the language `#1'. Using the pattern for}%
\typeout{** the default language instead.}%
\else
\language=\csname l@#1\endcsname
\fi
#2}}
\providecommand{\BIBdecl}{\relax}
\BIBdecl

\bibitem{1}
S.~E. Nicolò~A, Massaroni~C and S.~M., ``The importance of respiratory rate
  monitoring: From healthcare to sport and exercise,'' \emph{Sensors (Basel)}.

\bibitem{2}
M.~Tipton, A.~Harper, J.~Paton, and J.~Costello, ``The human ventilatory
  response to stress: Rate or depth?'' \emph{The Journal of Physiology}, vol.
  595, 06 2017.

\bibitem{3}
M.~Schneider, M.~Kraemmer, B.~Weber, and A.~Schwerdtfeger, ``Life events are
  associated with elevated heart rate and reduced heart complexity to acute
  psychological stress,'' \emph{Biological Psychology}, vol. 163, p. 108116, 05
  2021.

\bibitem{4}
M.~Szymkowski, D.~Hemmerling, M.~Stroiński, K.~Kwarciak, W.~Frier,
  O.~Georgiou, and M.~Maksymenko, \emph{Mid-air Haptic Biosignal Transfer}, 05
  2022, pp. 412--415.

\bibitem{19}
T.~Romanus, S.~Frish, M.~Maksymenko, W.~Frier, L.~Corenthy, and O.~Georgiou,
  ``Mid-air haptic bio-holograms in mixed reality,'' in \emph{2019 IEEE
  International Symposium on Mixed and Augmented Reality Adjunct
  (ISMAR-Adjunct)}, 2019, pp. 348--352.

\bibitem{5}
O.~Georgiou, H.~Limerick, L.~Corenthy, M.~Perry, M.~Maksymenko, S.~Frish,
  J.~M\"{u}ller, M.~Bachynskyi, and J.~R. Kim, ``Mid-air haptic interfaces for
  interactive digital signage and kiosks,'' in \emph{Extended Abstracts of the
  2019 CHI Conference on Human Factors in Computing Systems}.\hskip 1em plus
  0.5em minus 0.4em\relax New York, NY, USA: Association for Computing
  Machinery, 2019, p. 1–9.

\bibitem{6}
\BIBentryALTinterwordspacing
C.~Romano, E.~Schena, S.~Silvestri, and C.~Massaroni, ``Non-contact respiratory
  monitoring using an rgb camera for real-world applications,'' \emph{Sensors},
  vol.~21, no.~15, 2021. [Online]. Available:
  \url{https://www.mdpi.com/1424-8220/21/15/5126}
\BIBentrySTDinterwordspacing

\bibitem{7}
G.~Olani, ``Using video stream for continuous monitoring of breathing rate for
  general setting,'' \emph{Signal Image and Video Processing}, vol.~13, 10
  2019.

\bibitem{8}
\BIBentryALTinterwordspacing
C.~Massaroni, D.~Lo~Presti, D.~Formica, S.~Silvestri, and E.~Schena,
  ``Non-contact monitoring of breathing pattern and respiratory rate via rgb
  signal measurement,'' \emph{Sensors}, vol.~19, no.~12, 2019. [Online].
  Available: \url{https://www.mdpi.com/1424-8220/19/12/2758}
\BIBentrySTDinterwordspacing

\bibitem{9}
Y.~Nam, Y.~Kong, B.~Reyes, N.~Reljin, and K.~H. Chon, ``Monitoring of heart and
  breathing rates using dual cameras on a smartphone,'' \emph{PLOS ONE},
  vol.~11, pp. 1--15, 03 2016.

\bibitem{10}
B.~A. Reyes, N.~Reljin, Y.~Kong, Y.~Nam, and K.~H. Chon, ``Tidal volume and
  instantaneous respiration rate estimation using a volumetric surrogate signal
  acquired via a smartphone camera,'' \emph{IEEE Journal of Biomedical and
  Health Informatics}, vol.~21, no.~3, pp. 764--777, 2017.

\bibitem{11}
Y.~Lee, A.~Syakura, M.~Khalil, C.~Wu, Y.~Ding, and C.~Wang, ``A real-time
  camera-based adaptive breathing monitoring system,'' \emph{Medical and
  Biological Engineering and Computing}, vol.~59, no.~6, pp. 1285--1298, Jun.
  2021.

\bibitem{12}
\BIBentryALTinterwordspacing
C.~Lugaresi, J.~Tang, H.~Nash, C.~McClanahan, E.~Uboweja, M.~Hays, F.~Zhang,
  C.~Chang, M.~G. Yong, J.~Lee, W.~Chang, W.~Hua, M.~Georg, and M.~Grundmann,
  ``Mediapipe: {A} framework for building perception pipelines,'' \emph{CoRR},
  vol. abs/1906.08172, 2019. [Online]. Available:
  \url{http://arxiv.org/abs/1906.08172}
\BIBentrySTDinterwordspacing

\bibitem{13}
M.~Bartula, T.~Tigges, and J.~Muehlsteff, ``Camera-based system for contactless
  monitoring of respiration,'' in \emph{2013 35th Annual International
  Conference of the IEEE Engineering in Medicine and Biology Society (EMBC)},
  2013, pp. 2672--2675.

\bibitem{14}
W.~Wang and A.~C. den Brinker, ``Algorithmic insights of camera-based
  respiratory motion extraction,'' \emph{Physiological Measurement}, vol.~43,
  no.~7, p. 075004, jul 2022.

\bibitem{15}
J.~Canny, ``A computational approach to edge detection,'' \emph{IEEE
  Transactions on pattern analysis and machine intelligence}, no.~6, pp.
  679--698, 1986.

\bibitem{16}
S.~M. Guillaume~Heusch, André~Anjos, ``A reproducible study on remote heart
  rate measurement,'' 2016.

\bibitem{17}
Y.~Ren, B.~Syrnyk, and N.~Avadhanam, ``Dual attention network for heart rate
  and respiratory rate estimation,'' 10 2021.

\bibitem{18}
------, ``Improving video-based heart rate and respiratory rate estimation via
  pulse-respiration quotient,'' ser. Proceedings of Machine Learning Research,
  vol. 184.\hskip 1em plus 0.5em minus 0.4em\relax PMLR, 22 Jul 2022, pp.
  136--145.

\end{thebibliography}

\end{document}